\documentclass{article}
\usepackage{ijcai19}

\usepackage{times}
\usepackage{soul}
\usepackage{url}
\usepackage[hidelinks]{hyperref}
\usepackage[utf8]{inputenc}
\usepackage[small]{caption}
\usepackage{graphicx}
\usepackage{amsmath}
\usepackage{booktabs}
\usepackage{algorithm}
\usepackage{algorithmic}
\usepackage{subcaption}
\usepackage{multirow}

\usepackage{array}
\newcolumntype{L}[1]{>{\raggedright\let\newline\\\arraybackslash\hspace{0pt}}m{#1}}
\newcolumntype{C}[1]{>{\centering\let\newline\\\arraybackslash\hspace{0pt}}m{#1}}
\newcolumntype{R}[1]{>{\raggedleft\let\newline\\\arraybackslash\hspace{0pt}}m{#1}}

\hypersetup{
  colorlinks  = true, 
  citecolor   = [rgb]{0,0.08,0.45},  
  urlcolor   = [rgb]{0,0.08,0.45}   
}

\makeatletter
\def\blfootnote{\xdef\@thefnmark{}\@footnotetext}
\makeatother

\title{Artistic Enhancement and Style Transfer of Image Edges\\ using Directional Pseudo-coloring}

\author{
    Shouvik Mani
    \affiliations
    C3.ai \emails
    shouvik.mani@c3.ai
}

\begin{document}

\maketitle

\begin{abstract}
Computing the gradient of an image is a common step in computer vision pipelines. The image gradient quantifies the magnitude and direction of edges in an image and is used in creating features for downstream machine learning tasks. Typically, the image gradient is represented as a grayscale image. This paper introduces directional pseudo-coloring, an approach to color the image gradient in a deliberate and coherent manner. By pseudo-coloring the image gradient magnitude with the image gradient direction, we can enhance the visual quality of image edges and achieve an artistic transformation of the original image. Additionally, we present a simple style transfer pipeline which learns a color map from a style image and then applies that color map to color the edges of a content image through the directional pseudo-coloring technique. Code for the algorithms and experiments is available at \url{https://github.com/shouvikmani/edge-colorizer}.
\end{abstract}

\section{Introduction}

\blfootnote{This work was presented at the 2nd Workshop on Humanizing AI (HAI) at IJCAI'19 in Macao, China.}

Edge detection is a fundamental image processing technique which involves computing an image gradient to quantify the magnitude and direction of edges in an image. Image gradients are used in various downstream tasks in computer vision such as line detection, feature detection, and image classification. Despite their importance, image gradients are visually dull. They are typically represented as black-and-white or grayscale images visualizing the change in intensity at each pixel of the original image.

In this paper, we introduce directional pseudo-coloring, a technique to color the edges of an image (the image gradient) in a deliberate and coherent manner. We develop a pipeline to pseudo-color the image gradient magnitude using the image gradient direction, ensuring that edges are colored consistently according to their direction. An example of this transformation is shown in Figure 1.

The goal of this approach is to make image gradients more informative and appealing by enhancing them with color. A colored image gradient is necessary as the human eye can discern only two-dozen shades of gray, but thousands of shades of color~\cite{human_machine}. A colored image gradient will reveal latent information about edge direction not visible in a black-and-white gradient. In addition to its functional value, a colored image gradient will offer an aesthetically pleasing rendering of the original image.

\begin{figure}[t]
\begin{center}
\includegraphics[width=1.0\linewidth]{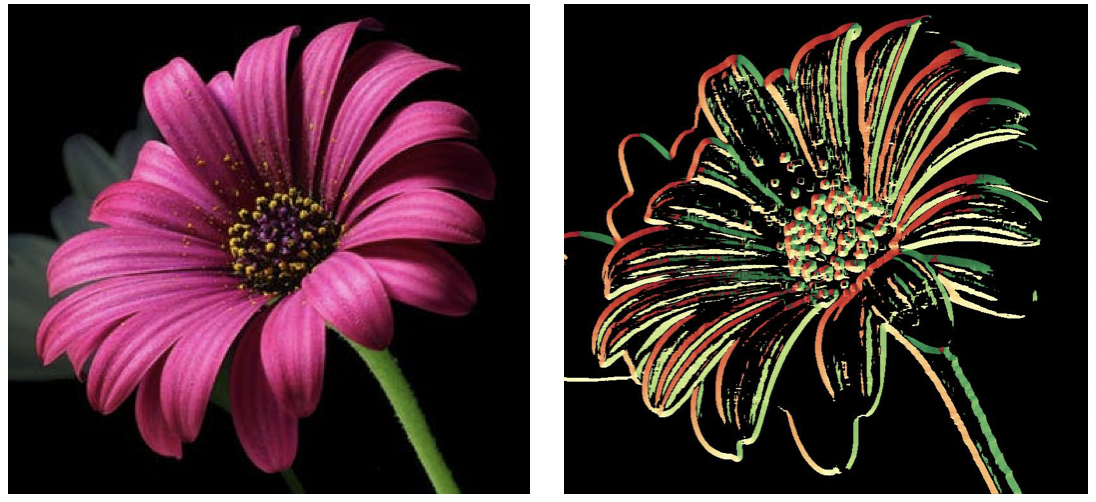}
\end{center}
\caption{An example of directional pseudo-coloring of image edges. Left: The original image. Right: Edge coloring result. }
\end{figure}

We also present a simple style transfer approach to give users more control over how they color image edges. This approach uses techniques from color quantization to learn the dominant colors in a style image. Then, it uses the directional pseudo-coloring technique to color the edges of a content image using the learned dominant colors.

\section{Related Work}

\begin{figure*}[h]
\centering
\includegraphics[height=6cm]{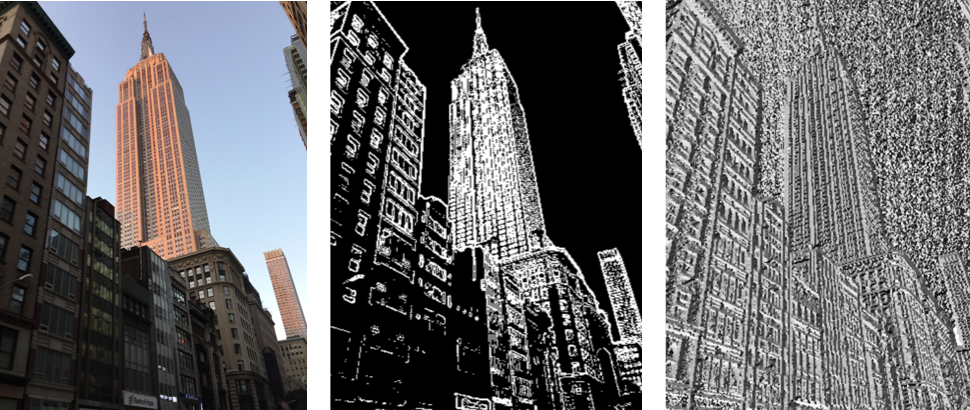}
\caption{Edge detection using Sobel filters. We convolve the original image (left) with the Sobel filters and compute the magnitude $G$ (middle) and direction $\theta$ (right) of the image gradient.}
\end{figure*}

Edge detection is a widely studied topic with many different approaches. Common methods for edge detection include using Sobel filters and the Canny edge detection algorithm. Szeliski~\cite{cv_book} reviews the fundamentals of edge detection and provides a brief survey of edge detection algorithms. In this paper, we will use Sobel filters due to their effectiveness and simplicity.

Our proposed edge coloring approach is an example of colorization, the process of adding color to a grayscale image or video. Edge detection has often been used to improve colorization results, but coloring the edges themselves has not been explored in depth. For instance, Huang et al.~\cite{Huang} have found that adding edge information may prevent the colorization process from bleeding over region boundaries in an image. 

Style transfer is a recent area of work which involves applying the appearance of a style image to the semantic content of a content image. Gatys et al.~\cite{Gatys} use Convolutional Neural Networks in order to create rich image representations and provide a neural style transfer algorithm to separate and recombine the image content and style images. We will present a simpler style transfer pipeline, aimed at applying styles specifically for the edge coloring task.

\section{Methodology}

The proposed pipeline for coloring image edges has two key components: edge detection and edge coloring. We review edge detection using Sobel filters, introduce the directional pseudo-coloring technique, and present some results from using the technique.

\subsection{Edge Detection}

The standard approach for edge detection is to convolve an image with a derivative filter like the Sobel filter to differentiate the intensity of the image pixels with respect to the horizontal and vertical directions of the image. However, since differentiation is very sensitive to noise, it is useful to blur the image first in order to detect the most salient edges~\cite{filtering_lecture}. This can be achieved by convolving the original image $f$ with a blur filter, such as the Gaussian filter, to produce a blurred image $f_b$ (line 1). For the purposes of edge detection, we will first convert $f$ to a grayscale image before blurring it.

\begin{align}
f_b = f * \frac{1}{16} \begin{bmatrix} 
1 & \quad 2 & \quad 1 \\ 
2 & \quad 4 & \quad 2 \\ 
1 & \quad 2 & \quad 1  
\end{bmatrix}
\end{align}

Once the image has been blurred, we can perform edge detection using Sobel filters~\cite{sobel}. We convolve the blurred image $f_b$ with the horizontal and vertical Sobel filters, $S_x$ and $S_y$ respectively, to approximate the image derivatives in the horizontal and vertical directions, $\frac{\partial f_b}{\partial x}$ and $\frac{\partial f_b}{\partial y}$ respectively (lines 2 -- 4). $\frac{\partial f_b}{\partial x}$ has a strong response to vertical edges, and $\frac{\partial f_b}{\partial y}$ has a strong response to horizontal edges.

\begin{align}
S_x = \begin{bmatrix} 
1 & 0 & -1 \\ 
2 & 0 & -2 \\ 
1 & 0 & -1  
\end{bmatrix}
\quad S_y = \begin{bmatrix} 
1 &  2 & 1 \\ 
0 &  0 & 0 \\ 
-1 & -2 & -1  
\end{bmatrix}
\end{align}

\begin{align}
\frac{\partial f_b}{\partial x} &= f_b * S_x \\
\frac{\partial f_b}{\partial y} &= f_b * S_y
\end{align}

$\frac{\partial f_b}{\partial x}$ and $\frac{\partial f_b}{\partial y}$ can be combined to form the image gradient $\nabla f_b$. Lines 5 -- 7 show how to compute the magnitude $G$ and direction $\theta$ of the image gradient.

\begin{align}
\nabla f_b &= \begin{bmatrix} \frac{\partial f_b}{\partial x}, \frac{\partial f_b}{\partial y} \end{bmatrix} \\
G &= \vert \vert \nabla f_b \vert \vert = \sqrt{\Big( \frac{\partial f_b}{\partial x} \Big)^2 + \Big( \frac{\partial f_b}{\partial y} \Big)^2} \\
\theta &= tan^{-1} \Big(\frac{\partial f_b}{\partial x} / \frac{\partial f_b}{\partial y}\Big)
\end{align}

$G$ and $\theta$ represent the intensity and direction of edges in the image respectively and are displayed in Figure 2. Not only are these images visually uninteresting, but they must also be viewed separately to understand both the intensity and direction of image edges. In the next section, we present an algorithm to color the gradient magnitude and unify the intensity and direction information of the edges into a single image. The quantities $G$ and $\theta$ will serve as inputs to our directional pseudo-coloring algorithm.

\subsection{Edge Coloring}

Once the edges have been detected, there are several possible ways to color them. One approach is to color the edges using the same colors from the original image, masking the thresholded gradient magnitude image over the original image. However, this does not add any novelty or aesthetic quality to the final image.

Instead, we choose to pseudo-color the gradient magnitude using values from the gradient direction. Pseudo-coloring is the process of mapping the shades of gray in a grayscale image to colors using a color map~\cite{pseudo_color}. We define a color map in lines 8 -- 10 as a function that maps normalized numerical values to RGB pixel values along some color spectrum~\cite{color_map}.

\begin{align}
color\_map: [0, 1] \rightarrow (R \in [0,255],\\ G \in [0,255],\\ B \in [0,255])
\end{align}

To implement our direction-based edge coloring approach, we create directional color maps, which map normalized direction values to RGB colors. Figure 3a provides examples of color maps in the Python Matplotlib library~\cite{Hunter} and Figure 3b illustrates the concept of a directional color map.

\begin{figure}[h]
\centering
\begin{subfigure}{0.4\textwidth}
\centering
\includegraphics[width=0.7\linewidth]{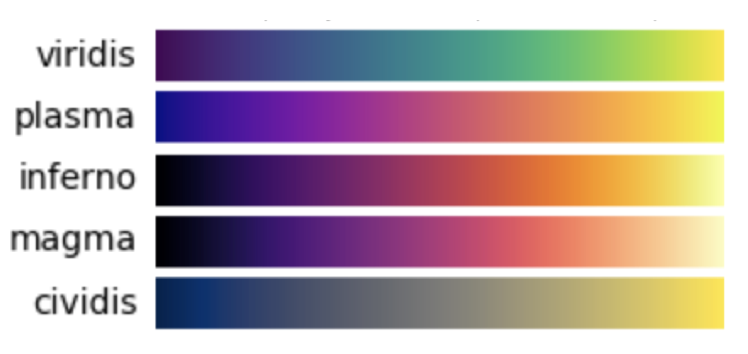} 
\caption{Examples of color maps in Matplotlib.}
\end{subfigure}
\begin{subfigure}{0.4\textwidth}
\centering
\includegraphics[width=0.7\linewidth]{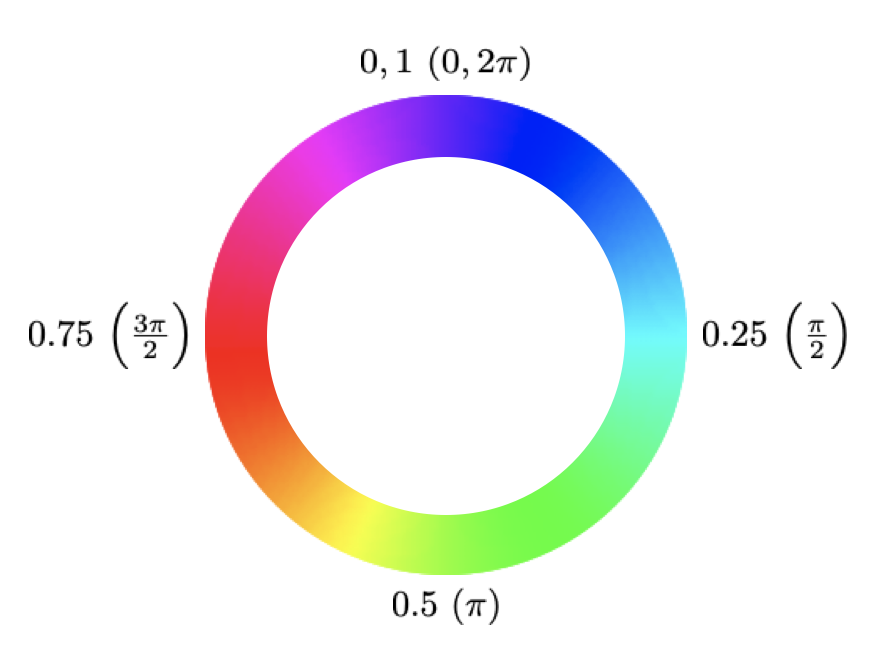}
\caption{A directional color map transforms normalized direction values to RGB colors.}
\end{subfigure}
\caption{The color map is a fundamental concept in directional pseudo-coloring.}
\end{figure}

After selecting a color map, we transform our gradient magnitude and direction to prepare them for coloring. First, in line 11, we threshold the gradient magnitude image, converting pixel $G_{x,y}$ to 1 if it has intensity above a threshold value $t$, and 0 otherwise. This thresholding turns the gradient magnitude into a black-and-white image $T$.

\begin{align}
T_{x,y} = \left\{
  \begin{array}{lr}
    1 \quad \text{if  } G_{x,y} \geq t \\
    0 \quad \text{otherwise}
  \end{array}
\right.
\end{align}

Then, in line 12, we mask the thresholded gradient magnitude over the gradient direction to create the colored image $C$. If $T_{x,y}$ is 0, then the pixel $C_{x, y}$ becomes a black pixel $(R=0, G=0, B=0)$. Otherwise, if $T_{x,y}$ is 1, we normalize the pixel's gradient direction $\theta_{x, y}$ between 0 and 1 and apply the chosen color map to color pixel $C_{x, y}$. Normalizing the gradient directions has the effect of creating a directional color map.

\begin{align}
C_{x,y} = \left\{
  \begin{array}{lr}
    color\_map \Big(\frac{\theta_{x,y} - min(\theta)}{max(\theta) - min(\theta)} \Big) \quad \text{if  } T_{x,y} = 1 \\
    (R=0, G=0, B=0) \quad \quad \quad \text{otherwise}
  \end{array}
\right.
\end{align}

The resulting image $C$ is an image of the thresholded gradient magnitude pseudo-colored by the normalized gradient direction. Because of this directional pseudo-coloring technique, edges along the same direction are assigned the same color, adding consistency and coherency to the image. The resulting image also has aesthetic value with edges of multiple colors contrasted against a dark background.

\section{Results}

\begin{figure*}[t]
\centering
\includegraphics[height=5cm]{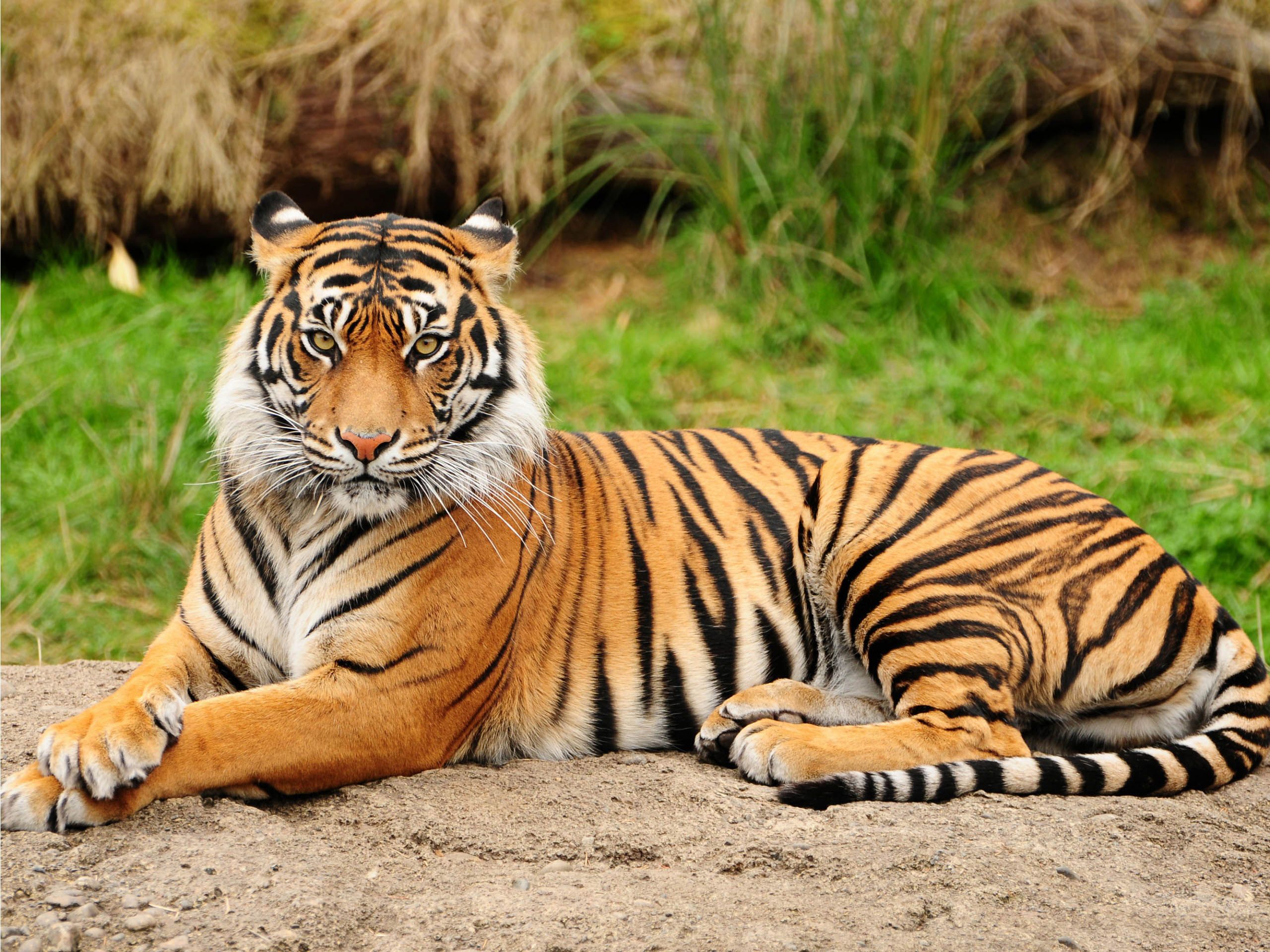}
\hspace{0.1cm}
\includegraphics[height=5cm]{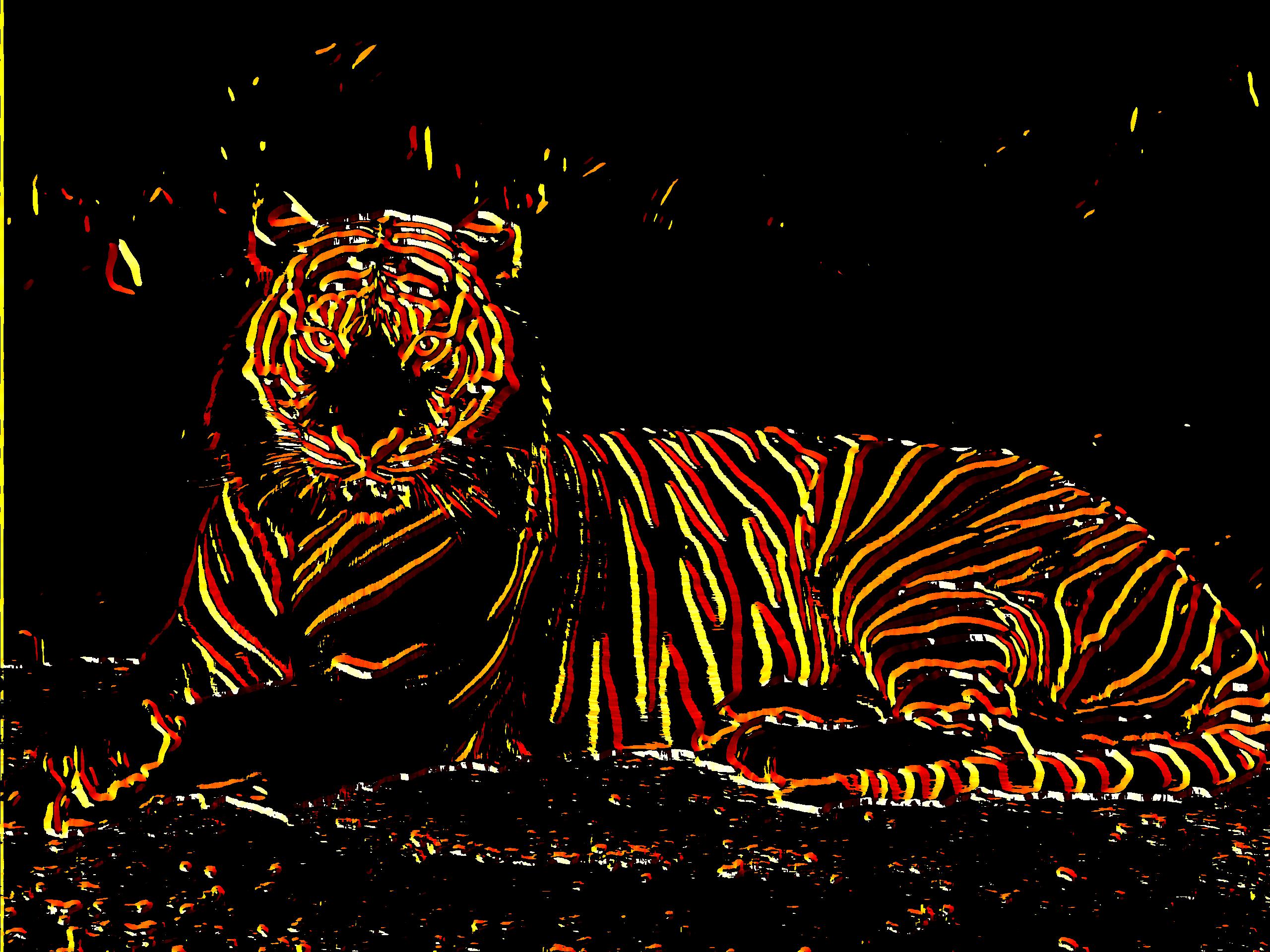}

\vspace{0.5cm}
\includegraphics[height=5cm]{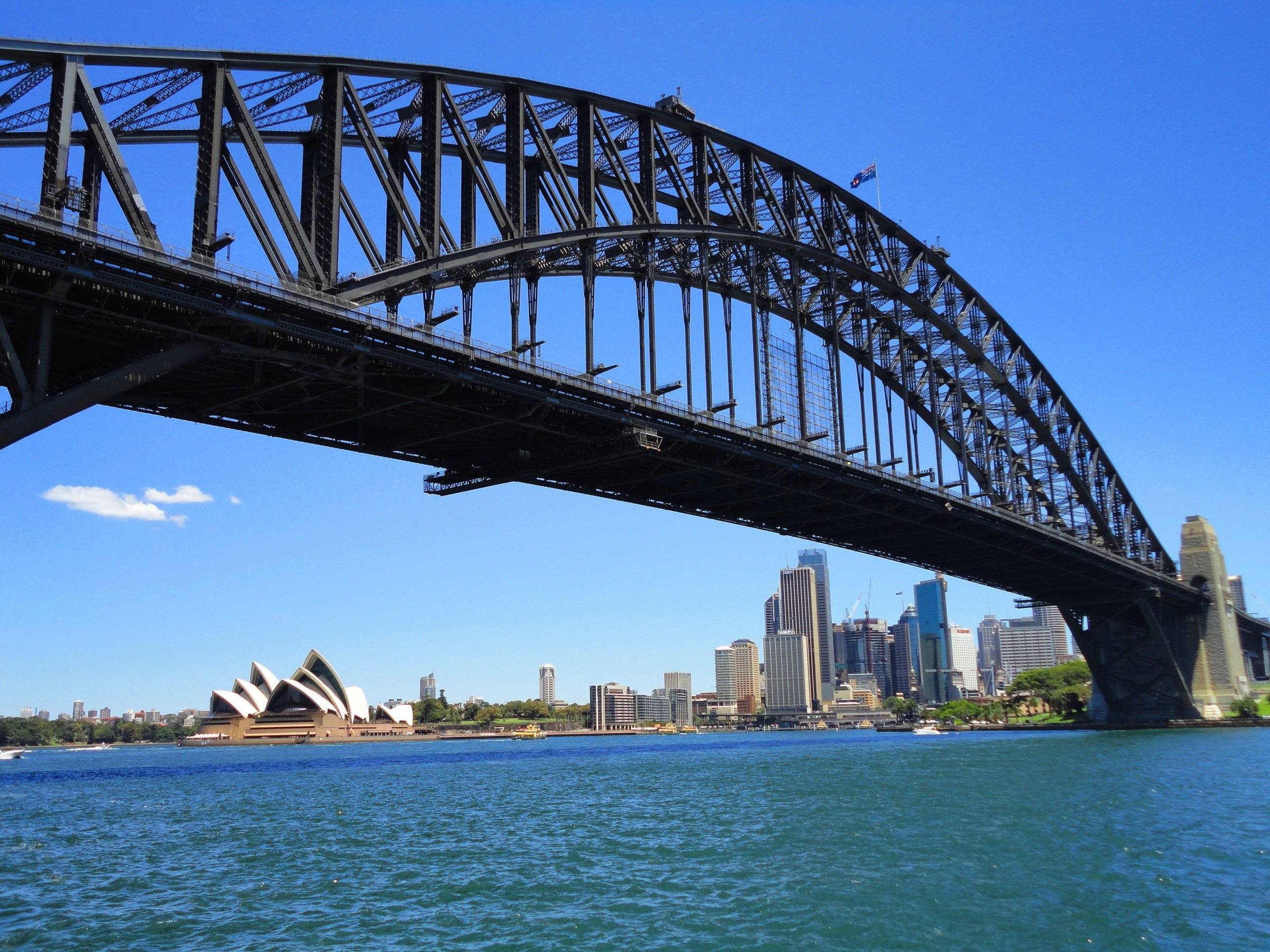}
\hspace{0.1cm}
\includegraphics[height=5cm]{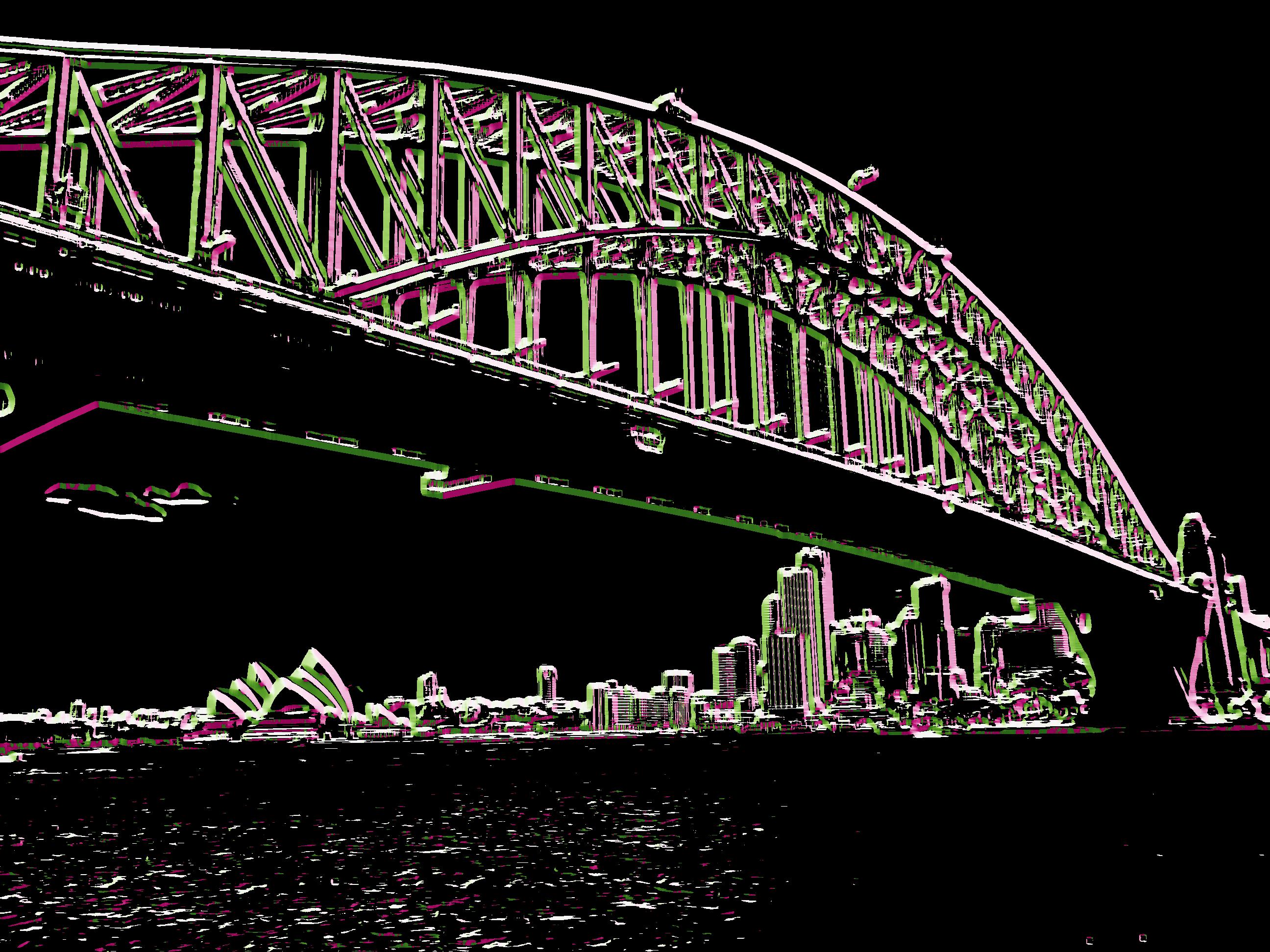}

\caption{Results from the edge coloring pipeline. Original images (left). Directional pseudo-colored edges (right).}
\end{figure*}

\begin{table*}[t!]
\centering
\begin{tabular}{ l  r  R{2.5cm}  R{2.5cm} }  
\toprule
\textbf{Edge Coloring Method}  &  \textbf{Type of Color Map}  &  \textbf{Mean NIMA Aesthetic Score}  &  \textbf{Mean NIMA Technical Score}  \\
\midrule
\multirow{2}{*}{\parbox{4cm}{Thresholded black \& white gradient magnitude}}    &   --         &   5.28    &   4.49  \\
\\
\midrule
\multirow{4}{*}{\parbox{4cm}{Directional pseudo-colored gradient magnitude}}                &   Sequential  &   5.24    &   4.61  \\
                                                  &   Diverging   &   5.24    &   4.67  \\
                                                  &   Cyclic      &   5.27    &   4.66  \\
                                                  &   Qualitative &   5.24    &   4.66  \\
\bottomrule
\end{tabular}
\caption{NIMA aesthetic and technical image quality scores for directional pseudo-coloring results using four types of color maps.}
\end{table*}

In Figure 4, we present a few examples of results from the edge coloring pipeline. The original images are shown on the left along with their directional pseudo-colored edges on the right. Color maps for the transformations are chosen from the set of sequential color maps in the Python Matplotlib library. The pseudo-colored edges are visually pleasing and detail-rich -- for instance, one can identify the individual beams and trusses on the bridge because edges in different directions are colored differently and consistently.

Apart from presenting these qualitative examples, we conduct a quantitative evaluation of our directional pseudo-coloring procedure. We devise an experiment to compare the aesthetic and technical qualities between the original black-and-white gradient magnitude $T$ and the pseudo-colored gradient magnitude $C$. While the aesthetic quality of an image pertains to its style and beauty, the technical quality measures pixel-level concepts such as noise, contrast, and saturation.

Although assessing the quality of an image is a highly subjective task, in recent years, researchers have made efforts to automate this process through machine learning. In our experiment, we leverage Neural Image Assessment (NIMA), a project which consists of two deep convolutional neural network (CNN) models which predict an image’s aesthetic and technical quality, respectively, on a scale of 1 -- 10~\cite{idealods2018imagequalityassessment,nima}. The two models are trained through transfer learning -- model weights are initialized as a CNN trained on an ImageNet object classification task and are fine-tuned to perform image quality assessment on the Aesthetic Visual Analysis (AVA) and Tampere Image Database (TID2013) datasets.

We use a collection of 100 high-resolution images from the DIVerse 2K Resolution (DIV2K) dataset for this experiment~\cite{DIV2K}. The diverse content and high-quality images in DIV2K makes it ideal for detecting and coloring edges. First, we perform edge detection and obtain the thresholded black-and-white gradient magnitude for each image in the dataset. Then, we pseudo-color the edges of each image using four random color maps from Matplotlib’s set of sequential, diverging, cyclic, and qualitative color maps (one random color map is chosen from each set).

Using the NIMA models, we predict the aesthetic and technical quality scores for the black-and-white gradient magnitude and for each of the four pseudo-colored gradient magnitudes, for each image. The mean NIMA scores for each type of color map are listed in Table 1. Regardless of the type of color map used, pseudo-coloring the edges does not improve the mean aesthetic score over the black-and-white edges, according to the NIMA model. However, the pseudo-colored edges have marginally higher mean technical scores compared to their black-and-white counterparts.

Overall, the NIMA scores suggest that the directional pseudo-coloring approach does not offer a significant improvement over standard black-and-white image gradients, contradicting our promising qualitative results. However, we should interpret these NIMA results with caution, as the NIMA models are not trained on assessing the quality of image gradients, but rather of natural images. More work is needed to quantify the factors that make image gradients attractive and useful.

\section{Application: Style Transfer}

\begin{figure*}[t]
\centering
\includegraphics[height=9cm]{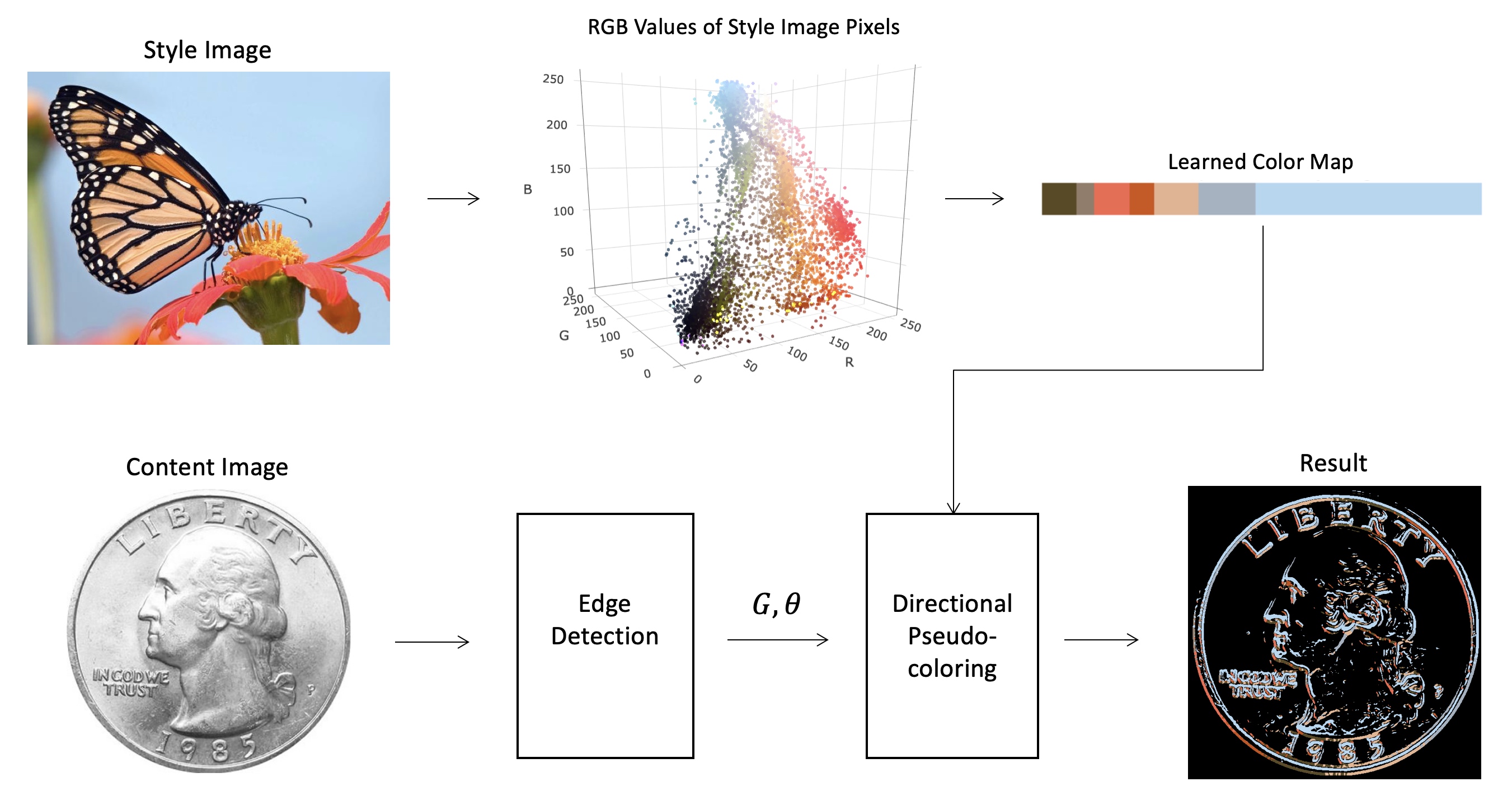}
\caption{Style transfer pipeline for edge coloring.}
\end{figure*}

In the examples shown so far, we have chosen input color maps for the directional pseudo-coloring algorithm arbitrarily. However, a user may wish to use a style image as input instead of an explicit color map, since a style image is much easier to specify. A natural extension of our edge coloring approach to meet this need is style transfer: specifying a style image to influence the coloring of edges in a content image. In this section, we develop a simple style transfer pipline to learn a color map from a style image and then apply that color map to color the edges of a content image.

Our style transfer approach draws inspiration from approaches to color quantization. As described by Orchard and Bouman~\cite{Orchard1991ColorQO}, color quantization is the problem of selecting an optimal color palette to reduce the number of colors in an image to a smaller, but representative set. While there are numerous algorithms for color quantization, Celebi~\cite{Celebi} demonstrates the effectiveness of the popular k-means clustering algorithm for this task. A tutorial on determining dominant colors in images through k-means clustering is provided by Tranberg~\cite{Tranberg}. We will use k-means clustering to identify dominant colors in the style image in order to form a color map for edge coloring of the content image.

We outline the style transfer pipeline in Figure 5. As input to the pipeline, the user must provide a content image, a style image, and the number of clusters $k$ (i.e. number of dominant colors) to learn from the style image. First, we represent each pixel in the style image as a point in the RGB color space. Using k-means clustering in this three-dimensional space, we identify $k$ clusters of pixels. The resulting cluster centers from the k-means algorithm represent the dominant colors in the style image.

After identifying the dominant colors, we form a color map using a greedy approach. To initialize the color map, we add a random dominant color as the first color in the color map. Then, until the remaining dominant colors are exhausted, we iteratively add the dominant color which is nearest (in terms of Euclidean distance in the RGB space) to the last added color in the color map. This greedy procedure helps create a sequential color map by ensuring that adjacent colors in the color map are similar to each other. A sequential color map is necessary for edges of the same direction to be assigned the same color in the edge coloring process.

Additionally, the size of each color in the color map is scaled by the number of pixels in the style image which are assigned to the cluster represented by that color. For instance, in Figure 5, the light blue color makes up the largest portion of the color map since most of the pixels in the style image are of that color. This scaling ensures that the learned color map is representative of the colors in the style image.

Once we have learned dominant colors from the style image through clustering and arranged those colors in a color map, we can proceed with edge detection and edge coloring for the content image. Now, we use the learned color map as an input to the directional pseudo-coloring algorithm. The resulting style transferred image preserves the edges of the content image, but is colored in with dominant colors from the style image. Figure 6 displays two examples of this style transfer process.

\begin{figure*}[t]
\centering
\includegraphics[height=9cm]{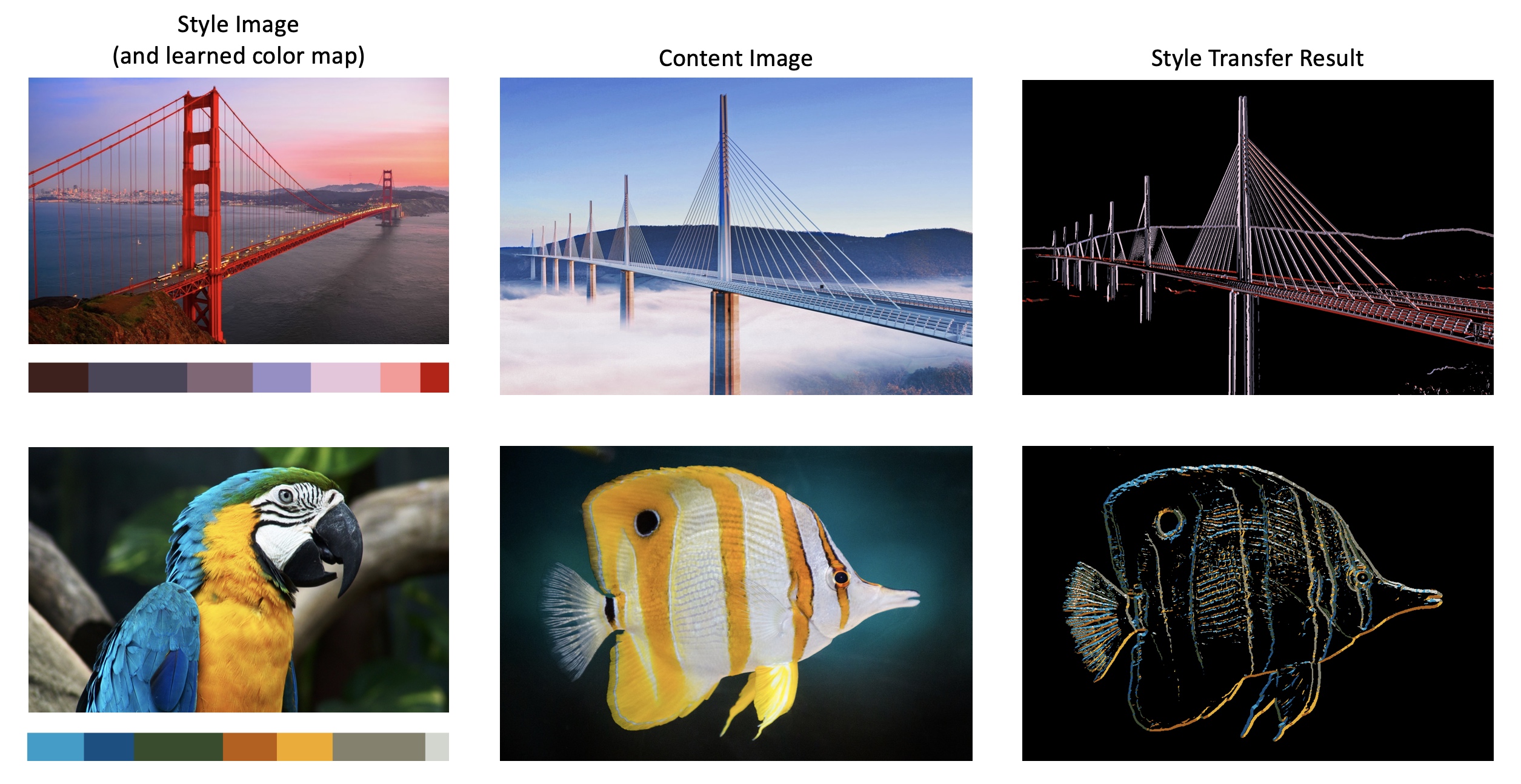}
\caption{Results from the style transfer pipeline.}
\end{figure*}

\section{Conclusion}

We have presented an algorithm to color the edges of an image in a way that adds both functional and aesthetic value. This approach leverages both the gradient magnitude and gradient direction to color image edges, producing a single visualization to understand the edges of an image. By enhancing image gradients with color, we can better visualize edges and create new artistic transformations of image edges.

To give users more flexibility in specifying colors for edge coloring, we have provided a style transfer pipeline which learns the dominant colors in a style image and colors the edges of a content image using those dominant colors. Although this approach is much simpler than deep learning-based style transfer techniques, it is limited to a transfer of dominant colors.

There is plenty of scope for further work to enhance and extend this edge coloring approach. First, one may wish to pseudo-color the gradient magnitude using a different quantity other than the gradient direction. For instance, pseudo-coloring the magnitude using the depth for each pixel (through depth estimation) may offer additional insights about the characteristics of the edges in an image. Second, we should evaluate how well this approach generalizes to three-dimensional figures: can a similar algorithm be used to create a 3D rendering of colored edges for a 3D figure? Finally, the style transfer approach may be improved by using other strategies for color quantization, such as by clustering the style image pixels in the HSV color space instead of the RGB color space.

\bibliographystyle{named}
\bibliography{enhancement_style_transfer_image_edges}

\end{document}